\documentclass[11pt]{article}
\usepackage{acl2014}
\usepackage{times}
\usepackage{latexsym}
\usepackage{graphicx} 
\usepackage{subfigure} 

\usepackage{amsmath}
\usepackage{amsfonts}
\usepackage{bm}
\usepackage[table]{xcolor}
\definecolor{lightgray}{gray}{0.9}
\definecolor{lessgray}{gray}{0.75}

\newcommand{\hQ}{\hat{Q}}
\newcommand{\hM}{\hat{M}}
\newcommand{\E}{\mathbb{E}}

\title{Low-dimensional Embeddings for Interpretable \\ Anchor-based Topic Inference}

\author{Moontae Lee\\
	    Dept. of Computer Science\\
	    Cornell University\\
	    Ithaca, NY, 14853\\
	    {\tt moontae@cs.cornell.edu}
	  \And
	David Mimno\\
  	Dept. of Information Science\\
  	Cornell University\\
  	Ithaca, NY, 14853\\
  {\tt mimno@cornell.edu}}

\date{}

\begin{document}
\maketitle

\begin{abstract}
The anchor words algorithm performs provably efficient topic model inference by finding an approximate convex hull in a high-dimensional word co-occurrence space. 
However, the existing greedy algorithm often selects poor anchor words, reducing topic quality and interpretability. 
Rather than finding an approximate convex hull in a high-dimensional space, we propose to find an exact convex hull in a visualizable 2- or 3-dimensional space. 
Such low-dimensional embeddings both improve topics and clearly show users why the algorithm selects certain words.
\end{abstract}

\section{Introduction}
\label{sect:intro}

Statistical topic modeling is useful in exploratory data analysis \cite{LDA}, but model inference is known to be NP-hard even for the simplest models with only two topics \cite{SR}, and training often remains a black box to users.
Likelihood-based training requires expensive approximate inference such as variational methods \cite{LDA}, which are deterministic but sensitive to initialization, or Markov chain Monte Carlo (MCMC) methods \cite{griffithsFinding}, which have no finite convergence guarantees.
Recently Arora et al. proposed the Anchor Words algorithm \cite{arora2013practical}, which casts topic inference as statistical recovery using a \textit{separability assumption}: each topic has a specific anchor word that appears only in the context of that single topic.
Each anchor word can be used as a unique pivot to disambiguate the corresponding topic distribution.
We then reconstruct the word co-occurrence pattern of each non-anchor words as a convex combination of the co-occurrence patterns of the anchor words. 

\begin{figure}[h]
\begin{center}
\hbox{\hspace{-0.65cm}\includegraphics[scale=0.185]{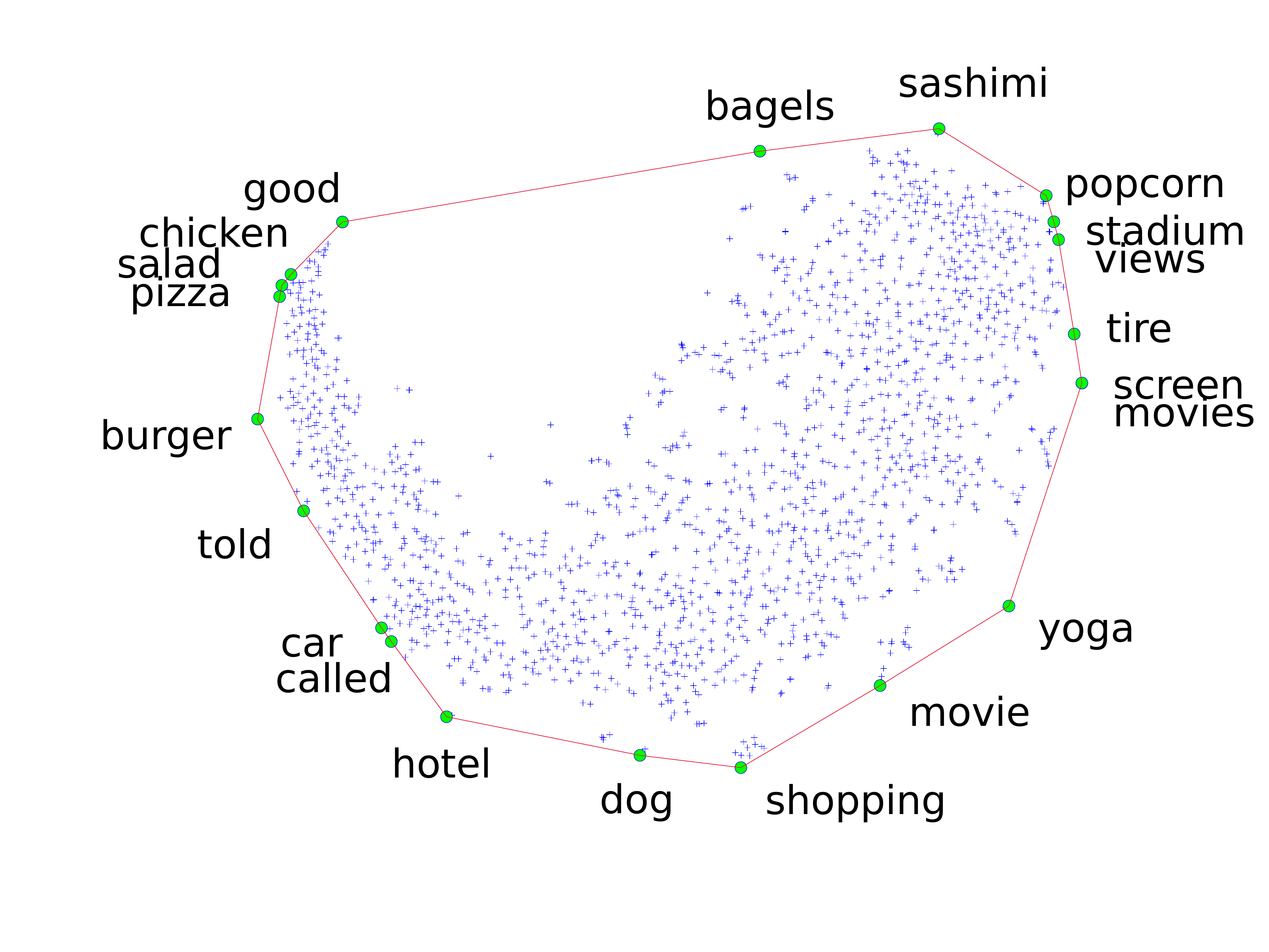}}
  \caption {2D $t$-SNE projection of a Yelp review corpus and its convex hull. The words corresponding to vertices are anchor words for topics, whereas non-anchor words correspond to the interior points.}
\label{fig:tsne-2d}
\end{center}
\end{figure}

This algorithm is fast, requiring only one pass through the training documents, and provides provable guarantees, but results depend entirely on selecting good anchor words.
\cite{arora2013practical} propose a greedy method that finds an approximate convex hull around a set of vectors corresponding to the word co-occurrence patterns for each vocabulary word.
Although this method is an improvement over previous work that used impractical linear programming methods \cite{AGM}, serious problems remain.
The method greedily chooses the farthest point from the current subspace until the given number of anchors have been found.
Particularly at the early stages of the algorithm, the words associated with the farthest points are likely to be infrequent and idiosyncratic, and thus form poor bases for human interpretation and topic recovery.
This poor choice of anchors noticeably affects topic quality: the anchor words algorithm tends to produce large numbers of nearly identical topics.

Besides providing a separability criterion, anchor words also have the potential to improve topic interpretability.
After learning topics for given text collections, users often request a label that summarizes each topic. 
Manually labeling topics is arduous, and labels often do not carry over between random initializations and models with differing numbers of topics. 
Moreover, it is hard to control the subjectivity in labelings between annotators, which is open to interpretive errors. 
There has been considerable interest in automating the labeling process \cite{mei2007automatic,lau2011automatic,chuang2012termite}.
\cite{chuang2012termite} propose a measure of {\em saliency}: a good summary term should be both distinctive specifically to one topic and probable in that topic.
Anchor words are by definition optimally distinct, and therefore may seem to be good candidates for topic labels, but greedily selecting extreme words often results in anchor words that have low probability.

In this work we explore the opposite of Arora et al.'s method: rather than finding an approximate convex hull for an exact set of vectors, we find an exact convex hull for an approximate set of vectors.
We project the $V \times V$ word co-occurrence matrix to visualizable 2- and 3-dimensional spaces using methods such as $t$-SNE \cite{vandermaaten2008visualizing}, resulting in an input matrix up to 3600 times narrower than the original input for our training corpora.
Despite this radically low-dimensional projection, the method not only finds topics that are as good or better than the greedy anchor method, it also finds highly salient, interpretable anchor words and provides users with a clear visual explanation for why the algorithm chooses particular words, all while maintaining the original algorithm's computational benefits.

\section{Related Work}

Latent Dirichlet allocation (LDA) \cite{LDA} models $D$ documents with a vocabulary $V$ using a predefined number of topics by $K$. 
LDA views both $\{A_k\}_{k=1}^K$, a set of $K$ topic-word distributions for each topic $k$, and $\{W_d\}_{d=1}^D$, a set of $D$ document-topic distributions for each document $d$, and $\{\mathbf{z}_d\}_{d=1}^{D}$, a set of topic-assignment vectors for word tokens in the document $d$, as randomly generated from known stochastic processes.
Merging $\{A_k\}$ as $k$-th column vector of $V \times K$ matrix $A$, $\{W_d\}$ as $d$-th column vector of $K \times D$ matrix $W$, the learning task is to estimate the posterior distribution of latent variables $A$, $W$, and $\{\mathbf{z}_d\}$ given $V \times D$ word-document matrix $\hM$, which is the only observed variable where $d$-th column corresponds to the empirical word frequencies in the training documents $d$.

\cite{arora2013practical} recover word-topic matrix $A$ and topic-topic matrix $R=\E[WW^T]$ instead of $W$ in the spirit of nonnegative matrix factorization.
Though the true underlying word distribution for each document is unknown and could be far from the sample observation $\hM$, 
the empirical word-word matrix $\hQ$ converges to its expectation $A\E[WW^T]A^T=ARA^T$ as the number of documents increases.
Thus the learning task is to approximately recover $A$ and $R$ pretending that the empirical $\hQ$ is close to the true second-order moment matrix $Q$.

The critical assumption for this method is to suppose that every topic $k$ has a specific anchor word $s_k$ that occurs with non-negligible probability ($> 0$) only in that topic.
The anchor word $s_k$ need not always appear in every document about the topic $k$, but we can be confident that the document is at least to some degree about the topic $k$  if it contains $s_k$.
This assumption drastically improves inference by guaranteeing the presence of a diagonal sub-matrix inside the word-topic matrix $A$.
After constructing an estimate $\hQ$, the algorithm in \cite{arora2013practical} first finds a set $S=\{s_1, ..., s_K\}$ of $K$ anchor words ($K$ is user-specified), and recovers $A$ and $R$ subsequently based on $S$. 
Due to this structure, overall performance depends heavily on the quality of anchor words.

In the matrix algebra literature this greedy anchor finding method is called \textit{QR with row-pivoting}.
Previous work classifies a matrix into two sets of row (or column) vectors where the vectors in one set can effectively reconstruct the vectors in another set, called \textit{subset-selection algorithms}. 
\cite{gu1996rrqr} suggest one important deterministic algorithm. A randomized algorithm provided by \cite{boutsidis2008cssp} is the state-of-the art using a pre-stage that selects the candidates in addition to \cite{gu1996rrqr}. 
We found no change in anchor selection using these algorithms, verifying the difficulty of the anchor finding process.
This difficulty is mostly because anchors must be nonnegative convex bases, whereas the classified vectors from the subset selection algorithms yield unconstrained bases.

The $t$-SNE model has previously been used to display high-dimensional embeddings of words in 2D space by Turian.\footnote{http://metaoptimize.com/projects/wordreprs/}
Low-dimensional embeddings of topic spaces have also been used to support user interaction with models:
\cite{eisenstein2011topicvis} use a visual display of a topic embedding to create a navigator interface.
Although $t$-SNE has been used to visualize the {\em results} of topic models, for example by \cite{lacoste2008disclda} and \cite{zhu2009medlda}, we are not aware of any use of the method as a fundamental component of topic inference.

\section{Low-dimensional Embeddings}

Real text corpora typically involve vocabularies in the tens of thousands of distinct words.
As the input matrix $\hQ$ scales quadratically with $V$, the Anchor Words algorithm must depend on a low-dimensional projection of $\hQ$ in order to be practical.
Previous work \cite{arora2013practical} uses random projections via either Gaussian random matrices \cite{johnson1984extensions} or sparse random matrices \cite{achlioptas2001database}, reducing the representation of each word to around 1,000 dimensions. 
Since the dimensionality of the compressed word co-occurrence space is an order of magnitude larger than $K$, we must still approximate the convex hull by choosing extreme points as before.

In this work we explore two projection methods: PCA and $t$-SNE \cite{vandermaaten2008visualizing}.
Principle Component Analysis (PCA) is a commonly-used dimensionality reduction scheme that linearly transforms the data to new coordinates where the largest variances are orthogonally captured for each dimension. By choosing only a few such principle axes, we can represent the data in a lower dimensional space.
In contrast, $t$-SNE embedding performs a non-linear dimensionality reduction preserving the local structures.
Given a set of points $\{\mathbf{x}_i\}$ in a high-dimensional space $X$, $t$-SNE allocates probability mass for each pair of points so that pairs of similar (closer) points become more likely to co-occur than dissimilar (distant) points. 
\begin{align}
  p_{j | i} &= \frac{\exp (-d(\mathbf{x}_i, \mathbf{x}_j)^2 / 2\sigma_i^2)}{\sum_{k \neq i} \exp (-d(\mathbf{x}_i, \mathbf{x}_k)^2 / 2\sigma_i^2)}\\
  p_{ij} &= \frac{p_{j | i} + p_{i | j}}{2N} \;\;\; \text{(symmetrized)}
\end{align}
\begin{figure}[tb]
\begin{center}
\hbox{\hspace{-0.3cm}\includegraphics[scale=0.185]{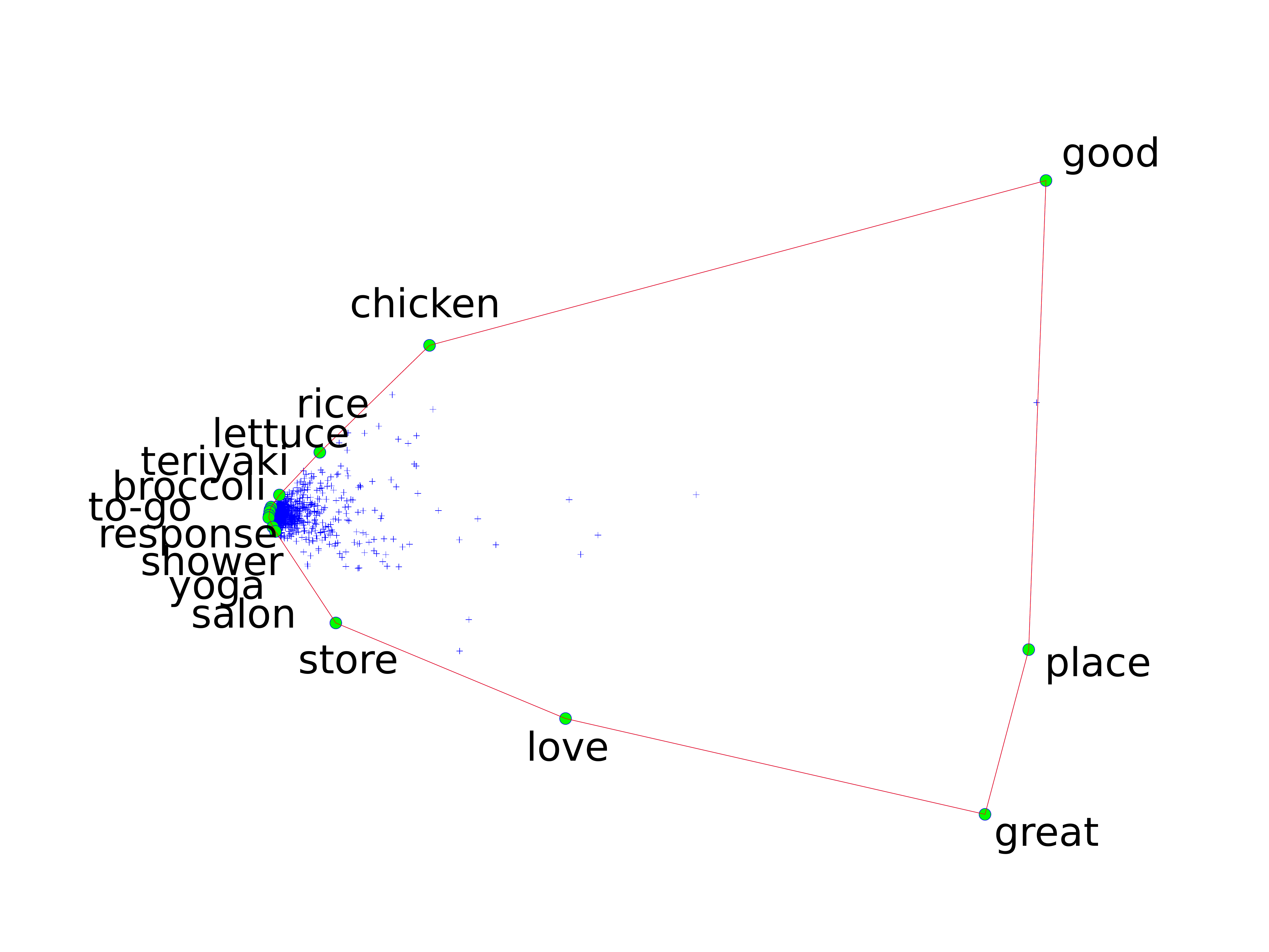}}
  \caption {2D PCA projections of a Yelp review corpus and its convex hulls.}
\label{fig:pca-2d}
\end{center}
\end{figure}

Then it generates a set of new points $\{\mathbf{y}_i\}$ in low-dimensional space $Y$ so that probability distribution over points in $Y$ behaves similarly to the distribution over points in $X$ by minimizing KL-divergence between two distributions:
\begin{align}
  q_{ij} = \frac{(1 + \|\mathbf{y}_i - \mathbf{y}_j \|^2)^{-1}}{\sum_{k \neq l} (1 + \|\mathbf{y}_k - \mathbf{y}_l \|^2)^{-1}}\\
  min \; KL(P || Q) = \sum_{i \neq j} p_{ij} \log \frac{p_{ij}}{q_{ij}}
\end{align}  

\begin{figure}[htb]
\begin{center}
\hspace*{-0.8cm}
\includegraphics[scale=0.19]{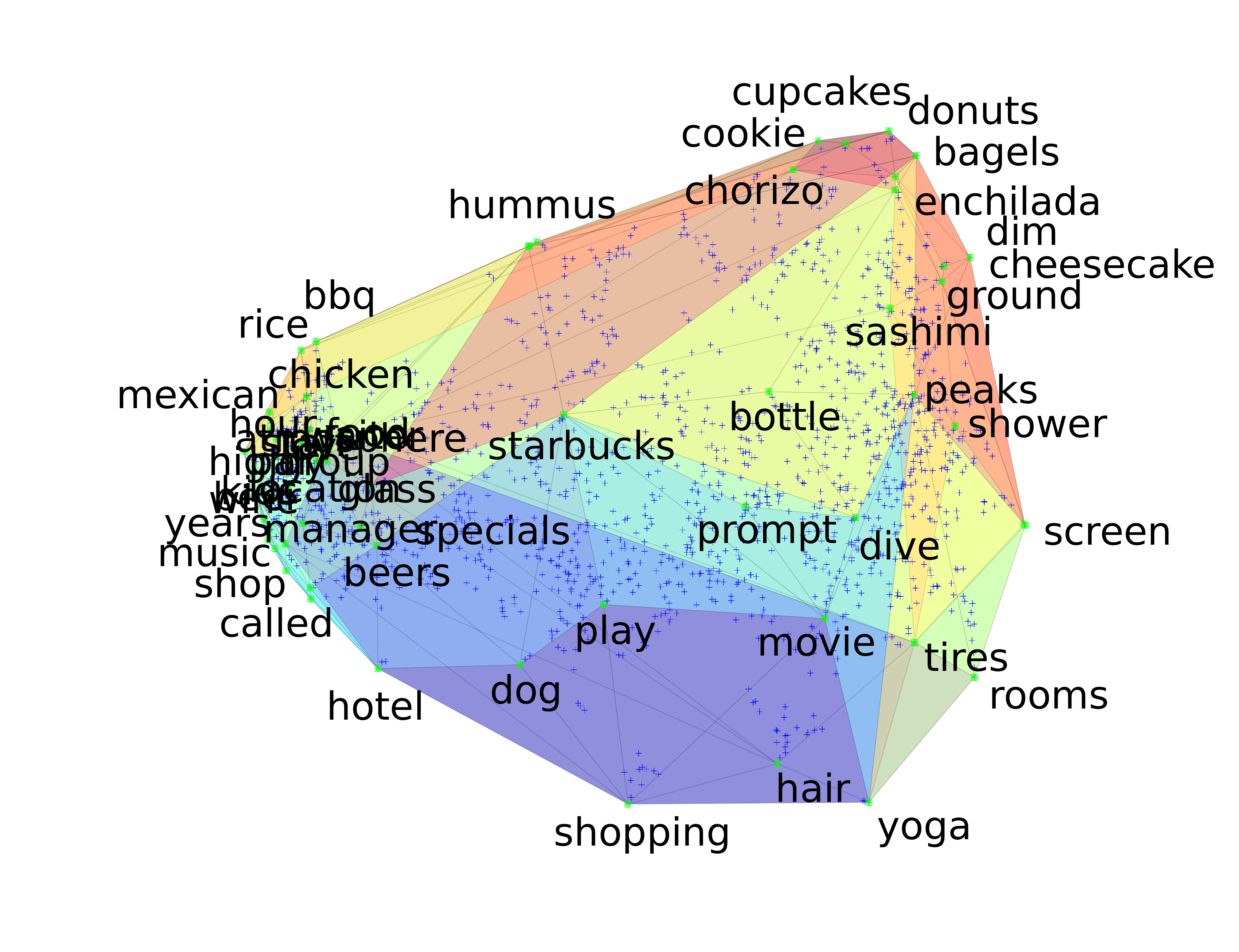}
  \caption {3D $t$-SNE projection of a Yelp review corpus and its convex hull. Vertices on the convex hull correspond to anchor words.}
\label{fig:tsne-3d}
\end{center}
\end{figure}

Instead of approximating a convex hull in such a high-dimensional space, we select the exact vertices of the convex hull formed in a low-dimensional projected space, which can be calculated efficiently. Figures \ref{fig:tsne-2d} and \ref{fig:pca-2d} show the convex hulls for 2D projections of $\hQ$ using  $t$-SNE and PCA for a corpus of Yelp reviews. Figure \ref{fig:tsne-3d} illustrates the convex hulls for 3D $t$-SNE projection for the same corpus. Anchor words correspond to the vertices of these convex hulls. 
Note that we present the 2D projections as illustrative examples only; we find that three dimensional projections perform better in practice.

In addition to the computational advantages, this approach benefits anchor-based topic modeling in two aspects. 
First, as we now compute the exact convex hull, the number of topics depends on the dimensionality of the embedding, $v$. 
For example in the figures, 2D projection has 21 vertices, whereas 3D projection supports 69 vertices. 
This implies users can easily tune the granularity of topic clusters by varying $v=2, 3, 4, ...$ without increasing the number of topics by one each time. 
Second, we can effectively visualize the thematic relationships between topic anchors and the rest of words in the vocabulary, enhancing both interpretability and options for further vocabulary curation.


\section{Experimental Results}

We find that radically low-dimensional $t$-SNE projections are effective at finding anchor words that are much more salient than the greedy method, and topics that are more distinctive, while maintaining comparable held-out likelihood and semantic coherence.
As noted in Section \ref{sect:intro}, the previous greedy anchor words algorithm tends to produce many nearly identical topics.
For example, 37 out of 100 topics trained on a 2008 political blog corpus have {\em obama, mccain, bush, iraq} or {\em palin} as their most probable word, including 17 just for {\em obama}.
Only 66\% of topics have a unique top word.
In contrast, the $t$-SNE model on the same dataset has only one topic whose most probable word is {\em obama}, and 86\% of topics have a unique top word ({\em mccain} is the most frequent top word, with five topics).

We use three real datasets: business reviews from the Yelp Academic Dataset,\footnote{https://www.yelp.com/academic\_dataset} political blogs from the 2008 US presidential election \cite{eisenstein2010cmu}, and New York Times articles from 2007.\footnote{http://catalog.ldc.upenn.edu/LDC2008T19}
Details are shown in Table \ref{tbl:corpus-stats}.
Documents with fewer than 10 word tokens are discarded due to possible instability in constructing $\hQ$.
We perform minimal vocabulary curation, eliminating a standard list of English stopwords\footnote{We used the list of 524 stop words included in the Mallet library.} and terms that occur below frequency cutoffs: 100 times (Yelp, Blog) and 150 times (NYT).
We further restrict possible anchor words to words that occur in more than three documents.
As our datasets are not artificially synthesized, we reserve 5\% from each set of documents for held-out likelihood computation.

\begin{table}[htp]
\begin{center}
{\small
\begin{tabular}{cccc}
Name & Documents & Vocab. & Avg. length \\
\hline
Yelp & 20,000 & 1,606 & 40.6 \\
Blog & 13,000 & 4,447 & 161.3 \\
NYT & 41,000 & 10,713 & 277.8 
\end{tabular}
}
\end{center}
\caption{Statistics for datasets used in experiments}
\label{tbl:corpus-stats}
\end{table}

Unlike \cite{arora2013practical}, which presents results on synthetic datasets to compare performance across different {\em recovery} methods given increasing numbers of documents, we are are interested in comparing {\em anchor finding} methods, and are mainly concerned with semantic quality.
As a result, although we have conducted experiments on synthetic document collections,\footnote{None of the algorithms are particularly effective at finding synthetically introduced anchor words possibly because there are other candidates around anchor vertices that approximate the convex hull to almost the same degree.} we focus on real datasets for this work.
We also choose to compare only anchor finding algorithms, so we do not report comparisons to likelihood-based methods, which can be found in \cite{arora2013practical}.

For both PCA and $t$-SNE, we use three-dimensional embeddings across all experiments. 
This projection results in matrices that are 0.03\% as wide as the original $V\times V$ matrix for the NYT dataset.
Without low-dimensional embedding, each word is represented by a V-dimensional vector where only several terms are non-zero due to the sparse co-occurrence patterns.
Thus a vertex captured by the greedy anchor-finding method is likely to be one of many eccentric vertices in very high-dimensional space. 
In contrast, $t$-SNE creates an effective dense representation where a small number of pivotal vertices are more clearly visible, improving both performance and interpretability.

Note that since we can find an {\em exact} convex hull in these spaces,\footnote{In order to efficiently find an exact convex hull, we use the {\em Quickhull} algorithm.} there is an upper bound to the number of topics that can be found given a particular projection.
If more topics are desired, one can simply increase the dimensionality of the projection.
For the greedy algorithm we use sparse random projections with 1,000 dimensions with 5\% negative entries and 5\% positive entries.
PCA and $t$-SNE choose (49, 32, 47) and (69, 77, 107) anchors, respectively for each of three Yelp, Blog, and NYTimes datasets.

\subsection{Anchor-word Selection}

We begin by comparing the behavior of low-dimensional embeddings to the behavior of the standard greedy algorithm.
Table \ref{tbl:anchor-list} shows ordered lists of the first 12 anchor words selected by three algorithms: $t$-SNE embedding, PCA embedding, and the original greedy algorithm.
Anchor words selected by $t$-SNE ({\em police, business, court}) are more general than anchor words selected by the greedy algorithm ({\em cavalry, al-sadr, yiddish}).
Additional examples of anchor words and their associated topics are shown in Table \ref{tbl:examples} and discussed in Section \ref{sect:examples}.

\begin{table}[ht]
\begin{center}
{\small
\begin{tabular}{c  c  c  c}
\hline
\# & t-SNE & PCA & Greedy\\\hline
1 & police & beloved & cavalry\\
2 & bonds & york & biodiesel\\
3 & business & family & h/w\\
4 & day & loving & kingsley\\
5 & initial & late & mourners\\
6 & million & president & pcl\\
7 & article & people & carlisle\\
8 & wife & article & al-sadr\\
9 & site & funeral & kaye\\
10 & mother & million & abc's\\
11 & court & board & yiddish\\
12 & percent & percent & great-grandmother\\
\hline
\end{tabular}
}
\end{center}
\caption{The first 12 anchor words selected by three algorithms for the NYT corpus. }
\label{tbl:anchor-list}
\end{table}

\begin{table}[t]
\begin{center}
{\tiny
\begin{tabular}{p{0.53cm}  p{0.03cm}  p{0.03cm}  p{5.41cm} }
\hline
Type & \# & HR & Top Words {\bf (Yelp)} \\
\hline
\rowcolor{lessgray}t-SNE & 16 &  0 & \textbf{mexican} good service great eat restaurant authentic delicious\\
\rowcolor{lightgray}PCA & 15 &  0 & \textbf{mexican} authentic eat chinese don't restaurant fast salsa\\
Greedy & 34 & 35 & good great food place service restaurant it's mexican\\
\hline
\rowcolor{lessgray}t-SNE & 6 & 0 & \textbf{beer} selection good pizza great wings tap nice\\
\rowcolor{lightgray}PCA & 39 & 6 & wine beer selection nice list glass \textbf{wines} bar\\
Greedy & 99 & 11 & beer selection great happy place wine good bar\\
\hline
\rowcolor{lessgray}t-SNE & 3 & 0 & \textbf{prices} great good service selection price nice quality\\
\rowcolor{lightgray}PCA & 12 & 0 & \textbf{atmosphere} prices drinks friendly selection nice beer ambiance\\
Greedy & 34 & 35 & good great food place service restaurant it's mexican\\
\hline
\rowcolor{lessgray}t-SNE & 10 & 0 & \textbf{chicken} salad good lunch sauce ordered fried soup\\
\rowcolor{lightgray}PCA & 10 & 0 & \textbf{chicken} salad lunch fried pita time back sauce\\
Greedy & 69 & 12 & chicken rice sauce fried ordered i'm spicy soup\\
\hline\hline
Type & \# &  HR & Top Words  {\bf (Blog)}\\
\hline
\rowcolor{lessgray}t-SNE & 10 & 0 & \textbf{hillary} clinton campaign democratic bill party win race\\
\rowcolor{lightgray}PCA & 4 & 0 & \textbf{hillary} clinton campaign democratic party bill democrats vote\\
Greedy & 45 & 19 & obama hillary campaign clinton obama's barack it's democratic\\
\hline
\rowcolor{lessgray}t-SNE & 3 & 0 & \textbf{iraq} war troops iraqi mccain surge security american\\
\rowcolor{lightgray}PCA & 9 & 1 & iraq \textbf{iraqi} war troops military forces security american\\
Greedy & 91 & 8 & iraq mccain war bush troops withdrawal obama iraqi\\
\hline
\rowcolor{lessgray}t-SNE & 9 & 0 & \textbf{allah} muhammad qur verses unbelievers ibn muslims world\\
\rowcolor{lightgray}PCA & 18 & 14 & allah muhammad qur verses unbelievers story time update\\
Greedy & 4 & 5 & allah muhammad people qur verses \textbf{unbelievers} ibn sura\\
\hline
\rowcolor{lessgray}t-SNE & 19 & 0 & \textbf{catholic} abortion catholics life hagee time biden human\\
\rowcolor{lightgray}PCA & 2 & 0 & \textbf{people} it's time don't good make years palin\\
Greedy & 40 & 1 & abortion \textbf{parenthood} planned people time state life government\\
\hline\hline
Type & \# & HR & Top Words  {\bf (NYT)}\\
\hline
\rowcolor{lessgray}t-SNE & 0 & 0 & \textbf{police} man yesterday officers shot officer year-old charged\\
\rowcolor{lightgray}PCA & 6 & 0 & \textbf{people} it's police way those three back don't\\
Greedy & 50 & 198 & police man yesterday officers officer people street city\\
\hline
\rowcolor{lessgray}t-SNE & 19 & 0 & \textbf{senator} republican senate democratic democrat state bill \\
\rowcolor{lightgray}PCA & 33 & 2 & state republican \textbf{republicans} senate senator house bill party\\
Greedy & 85 & 33 & senator republican president state campaign presidential people \\
\hline
\rowcolor{lessgray}t-SNE & 2 & 0 & \textbf{business} chief companies executive group yesterday billion \\
\rowcolor{lightgray}PCA & 21 & 0 & \textbf{billion} companies business deal group chief states united\\
Greedy & 55 & 10 & radio business companies percent day music article satellite\\
\hline
\rowcolor{lessgray}t-SNE & 14 & 0 & \textbf{market} sales stock companies prices billion investors price\\
\rowcolor{lightgray}PCA & 11 & 0 & \textbf{percent} market rate week state those increase high\\
Greedy & 77 & 44 & companies percent billion million group business chrysler people\\
\hline
\end{tabular}
}
\end{center}
\caption{Example $t$-SNE topics and their most similar topics across algorithms. The Greedy algorithm  can find similar topics, but the anchor words are much less salient.}
\label{tbl:examples}
\end{table}

The Gram-Schimdt process used by Arora et al. greedily selects anchors in high-dimensional space.
As each word is represented within $V$-dimensions, finding the word that has the next most distinctive co-occurrence pattern tends to prefer overly eccentric words with only short, intense bursts of co-occurring words.
While the bases corresponding to these anchor words could be theoretically relevant for the original space in high-dimension, they are less likely to be equally important in low-dimensional space.
Thus projecting down to low-dimensional space can rearrange the points emphasizing not only  uniqueness, but also longevity, achieving the ability to form measurably more specific topics.

Concretely, neither {\em cavalry, al-sadr, yiddish} nor {\em police, business, court} are full representations of New York Times articles, but the latter is a much better basis than the former due to its greater generality.
We see the effect of this difference in the specificity of the resulting topics (for example in 17 {\em obama} topics).
Most words in the vocabulary have little connection to the word {\em cavalry}, so the probability $p(z|w)$ does not change much across different $w$. 
When we convert these distributions into $P(w|z)$ using the Bayes' rule, the resulting topics are very close to the corpus distribution, a unigram distribution $p(w)$.
\begin{align*}
\label{eq:cavalry}
  p(w|z = k_{cavalry}) &\propto p(z=k_{cavalry} | w)p(w) \\
  &\approx p(w)
\end{align*}
This lack of specificity results in the observed similarity of topics.

\subsection{Quantitative Results}
\label{sect:examples}
In this section we compare PCA and $t$-SNE projections to the greedy algorithm along several quantitative metrics.
To show the effect of different values of $K$, we report results for varying numbers of topics.
As the anchor finding algorithms are deterministic, the anchor words in a $K$-dimensional model are identical to the first $K$ anchor words in a $(K+1)$-dimensional model.
For the greedy algorithm we select anchor words in the order they are chosen.
For the PCA and $t$-SNE methods, which find anchors jointly, we sort words in descending order by their distance from their centroid.

\textbf{Recovery Error.}
Each non-anchor word is approximated by a convex combination of the $K$ anchor words.
The projected gradient algorithm \cite{arora2013practical} determines these convex coefficients so that the gap between the original word vector and the approximation becomes minimized. 
As choosing a good basis of $K$ anchor words decreases this gap, Recovery Error (RE) is defined by the average $\ell_2$-residuals across all words.
\begin{equation}
  RE = \frac{1}{V} \sum_{i = 1}^V \| \bar{Q}_i - \sum_{k=1}^K p(z_1 = k | w_1 = i) \bar{Q}_{S_k} \|_2
\end{equation}
\begin{figure}[htb]
\begin{center}
\hspace*{-0.27cm}
    \includegraphics[scale=0.55]{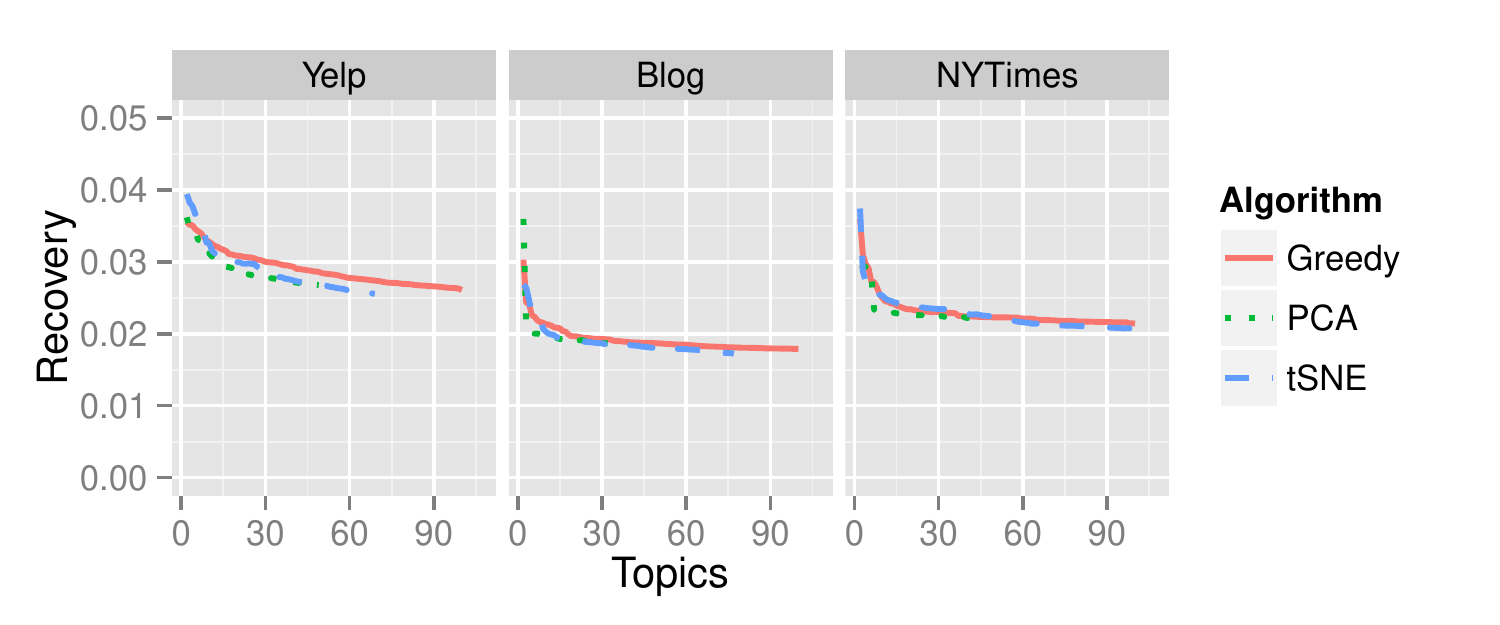}
    \caption {Recovery error is similar across algorithms.}
    \label{fig:recovery}
\end{center}
\end{figure}
Recovery error decreases with the number of topics, and improves substantially after the first 10--15 anchor words for all methods.
The $t$-SNE method has slightly better performance than the greedy algorithm, but they are similar.
Results for recovery with the original, unprojected matrix (not shown) are much worse than the other algorithms, suggesting that the initial anchor words chosen are especially likely to be uninformative.

\textbf{Normalized Entropy.}
As shown previously, if the probability of topics given a word is close to uniform, the probability of that word in topics will be close to the corpus distribution.
Normalized Entropy (NE) measures the entropy of this distribution relative to the entropy of a $K$-dimensional uniform distribution:
\begin{equation}
 NE = \frac{1}{V} \sum_{i=1}^V \frac{H(z | w = i)}{\log{K}}.
\end{equation}
\begin{figure}[htb]
\begin{center}
\hspace*{-0.20cm}
    \includegraphics[scale=0.55]{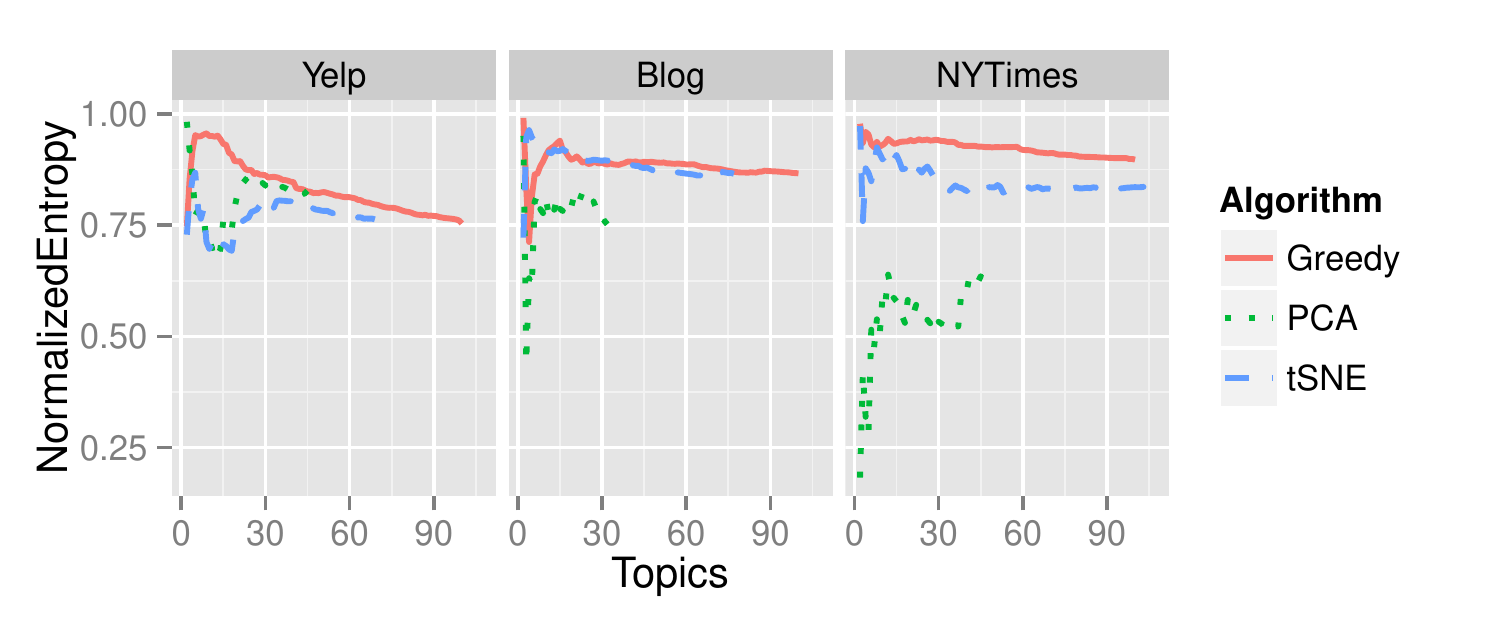}
    \caption {Words have higher topic entropy in the greedy model, especially in NYT, resulting in less specific topics.}
    \label{fig:entropy}
\end{center}
\end{figure}
The normalized entropy of topics given word distributions usually decreases as we add more topics, although both $t$-SNE and PCA show a dip in entropy for low numbers of topics.
This result indicates that words become more closely associated with particular topics as we increase the number of topics.
The low-dimensional embedding methods ($t$-SNE and PCA) have consistently lower entropy.

\textbf{Topic Specificity} and  \textbf{Topic Dissimilarity.}
We want topics to be both specific (that is, not overly general) and different from each other.
When there are insufficient number of topics, $p(w | z)$ often resembles the corpus distribution $p(w)$, where high frequency terms become the top words contributing to most topics.
Topic Specificity (TS) is defined by the average KL divergence from each topic's conditional distribution to the corpus distribution.\footnote{We prefer {\em specificity} to \cite{alsumait2009topic}'s term {\em vacuousness} because the metric increases as we move away from the corpus distribution.}
\begin{equation}
  TS = \frac{1}{K} \sum_{k=1}^K KL\big(p(w | z = k) \;||\; p(w)\big)
\end{equation}
One way to define the distance between multiple points is the minimum radius of a ball that covers every point. 
Whereas this is simply the distance form the centroid to the farthest point in the Euclidean space, it is an itself difficult optimization problem to find such centroid of distributions under metrics such as KL-divergence and Jensen-Shannon divergence.
To avoid this problem, we measure Topic Dissimilarity (TD) viewing each conditional distribution $p(w | z)$ as a simple $V$-dimensional vector in $\mathbb{R}^V$. 
Recall $a_{ik} = p(w = i | z = k)$,
\begin{equation}
  TD = \max_{1 \leq k \leq K} \| \frac{1}{K}\sum_{k'=1}^K a_{*k'} - a_{*k} \|_2 .
\end{equation}
\begin{figure}[htb]
\begin{center}
\hspace*{-0.27cm}
    \includegraphics[scale=0.55]{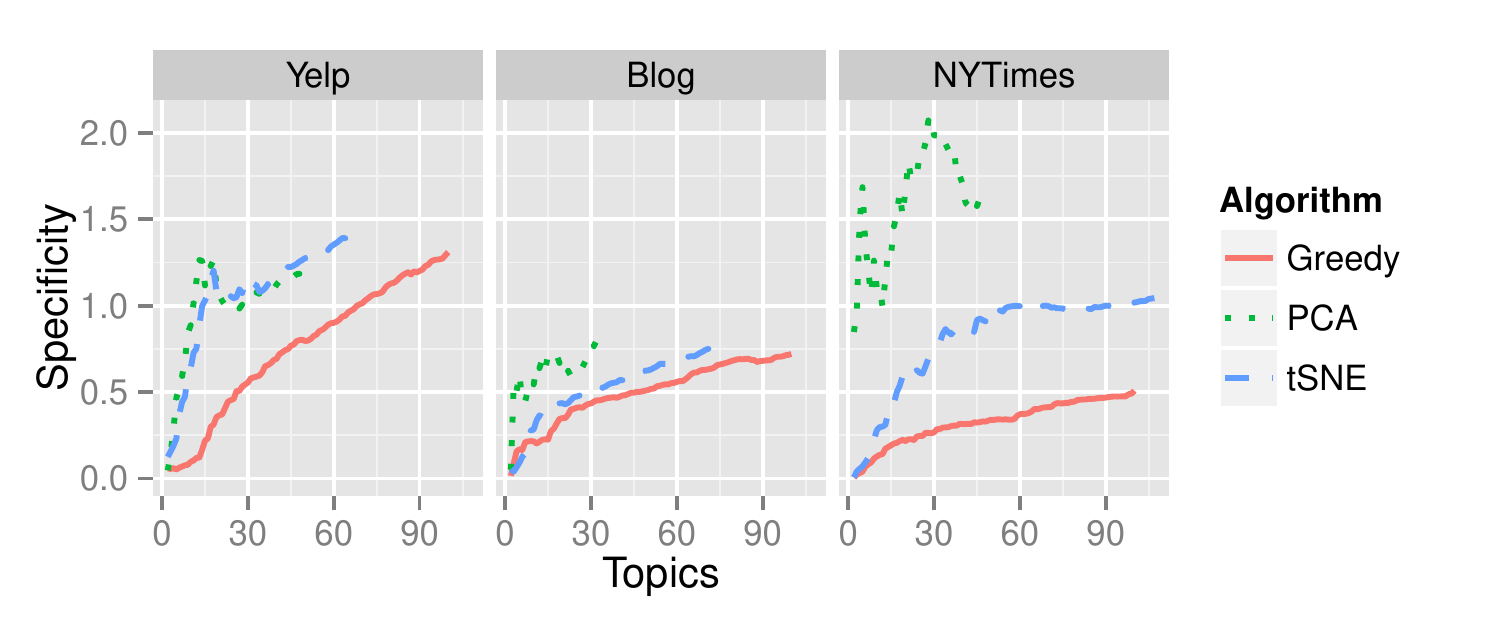}
\hspace*{-0.27cm}
    \includegraphics[scale=0.55]{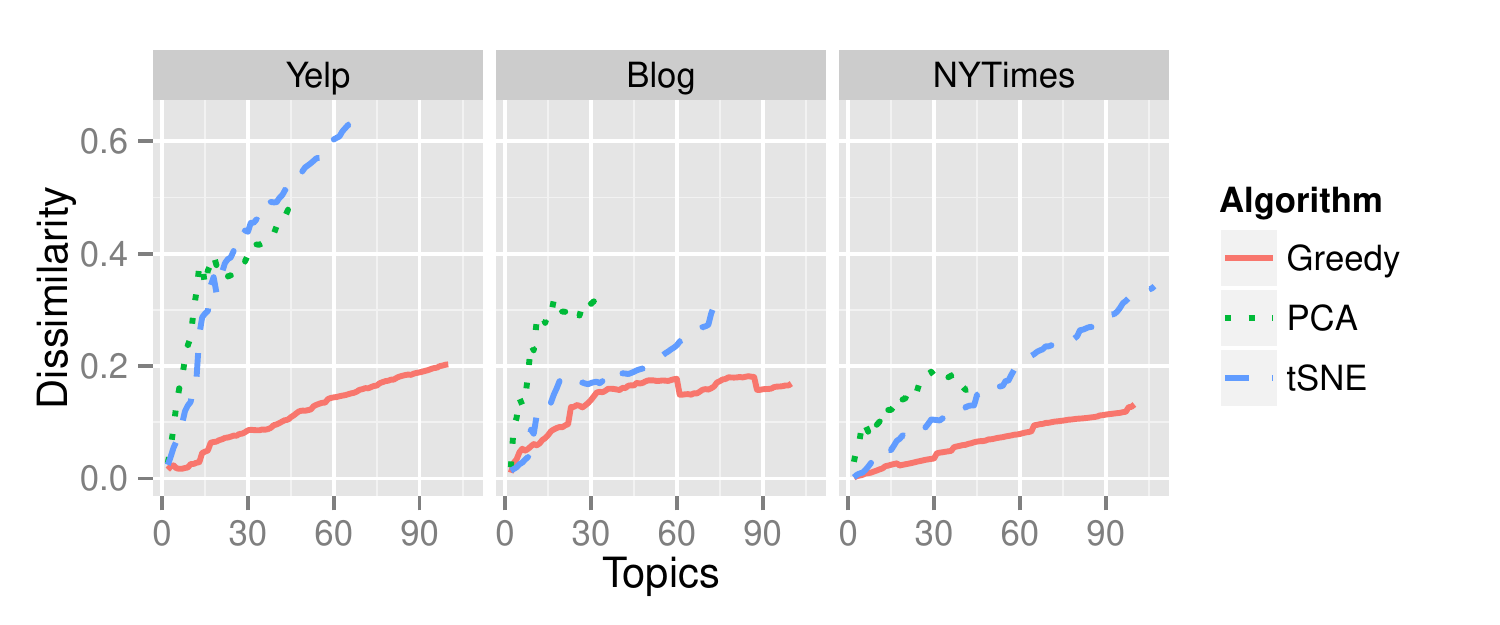}
    \caption {Greedy topics look more like the corpus distribution and more like each other.}
    \label{fig:spec-dis}
\end{center}
\end{figure}
Specificity and dissimilarity increase with the number of topics, suggesting that with few  anchor words, the topic distributions are close to the overall corpus distribution and very similar to one another.
The $t$-SNE and PCA algorithms produce consistently better specificity and dissimilarity, indicating that they produce more useful topics early with small numbers of topics.
The greedy algorithm produces topics that are closer to the corpus distribution and less distinct from each other (17 {\em obama} topics).

\textbf{Topic Coherence} is known to correlate with the semantic quality of topic judged by human annotators \cite{mimno2011optimizing}.
Let $\mathcal{W}_k^{(T)}$ be $T$ most probable words (i.e., top words) for the topic $k$. 
\begin{equation}
  TC = \sum_{w_1 \neq w_2 \in \mathcal{W}_k^{(T)}} \log \frac{D(w_1, w_2) + \epsilon}{D(w_1)}
\end{equation}
Here $D(w_1, w_2)$ is the co-document frequency, which is the number of documents in $\mathcal{D}$ consisting of two words $w_1$ and $w_2$ simultaneously.
$D(w)$ is the simple document frequency with the word $w$. 
The numerator contains smoothing count $\epsilon$ in order to avoid taking the logarithm of zero.
\begin{figure}[htb]
\begin{center}
  \hspace*{-0.20cm}
    \includegraphics[scale=0.55]{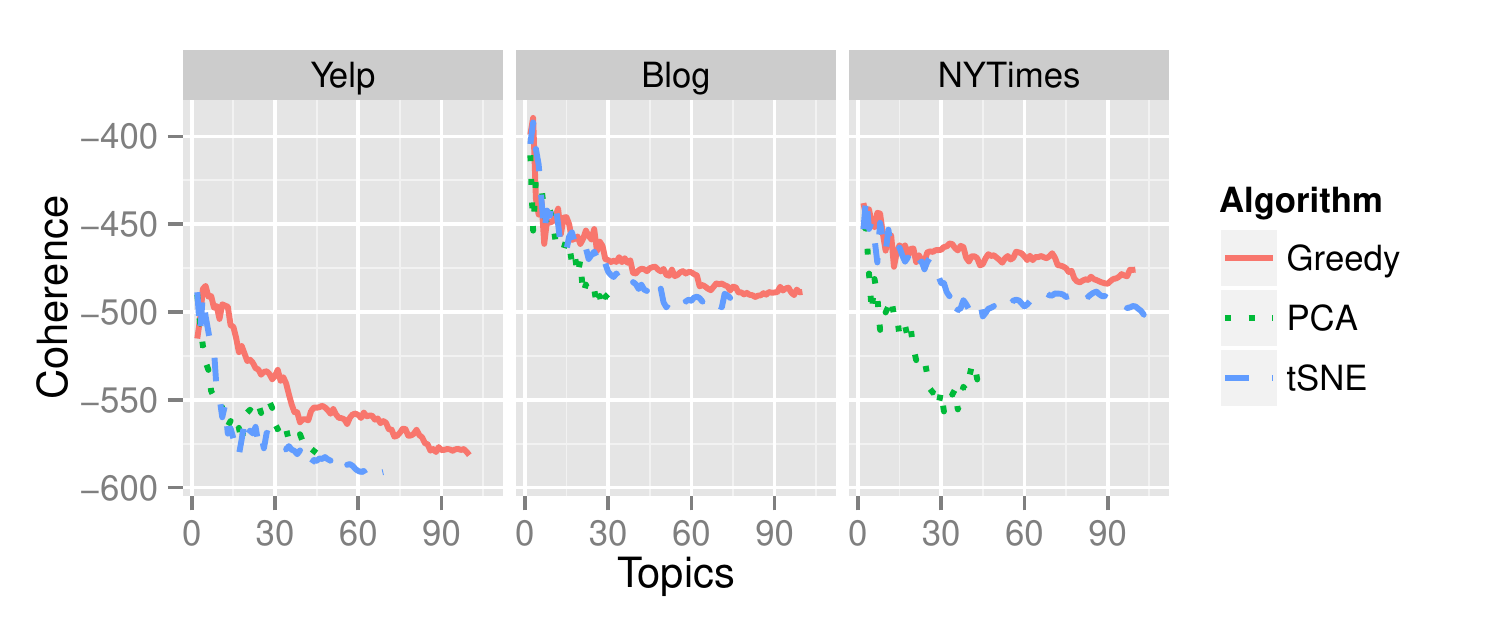}
    \caption {The greedy algorithm creates more coherent topics (higher is better), but at the cost of many overly general or repetitive topics.}
    \label{fig:coherence}
\end{center}
\end{figure}
Coherence scores for $t$-SNE and PCA are worse than those for the greedy method, but this result must be understood in combination with the Specificity and Dissimilarity metrics.
The most frequent terms in the overall corpus distribution $p(w)$ often appear together in documents.
Thus a model creating many topics similar to the corpus distribution is likely to achieve high Coherence, but low Specificity by definition.

\textbf{Saliency}. \cite{chuang2012termite} define saliency for topic words as a combination of distinctiveness and probability within a topic.
Anchor words are distinctive by construction, so we can increase saliency by selecting more probable anchor words.
We measure the probability of anchor words in two ways.
First, we report the zero-based rank of anchor words within their topics.
Examples of this metric, which we call ``hard'' rank are shown in Table \ref{tbl:examples}.
The hard  rank  of the anchors in the PCA and $t$-SNE models are close to zero, while the anchor words for the greedy algorithm are much lower ranked, well below the range usually displayed to users.
Second, 
while hard rank measures the perceived difference in rank of contributing words, position may not fully capture the relative importance of the anchor word. ``Soft'' rank quantifies the average log ratio between probabilities of the prominent word $w_k^*$ and the anchor word $s_k$.
\begin{equation}
  SR = \frac{1}{K} \sum_{k=1}^K \log \frac{p(w = w_k^* | z = k)}{p(w = s_k | z = k)}
\end{equation}

\begin{figure}[htb]
\begin{center}
\hspace*{-0.27cm}
    \includegraphics[scale=0.55]{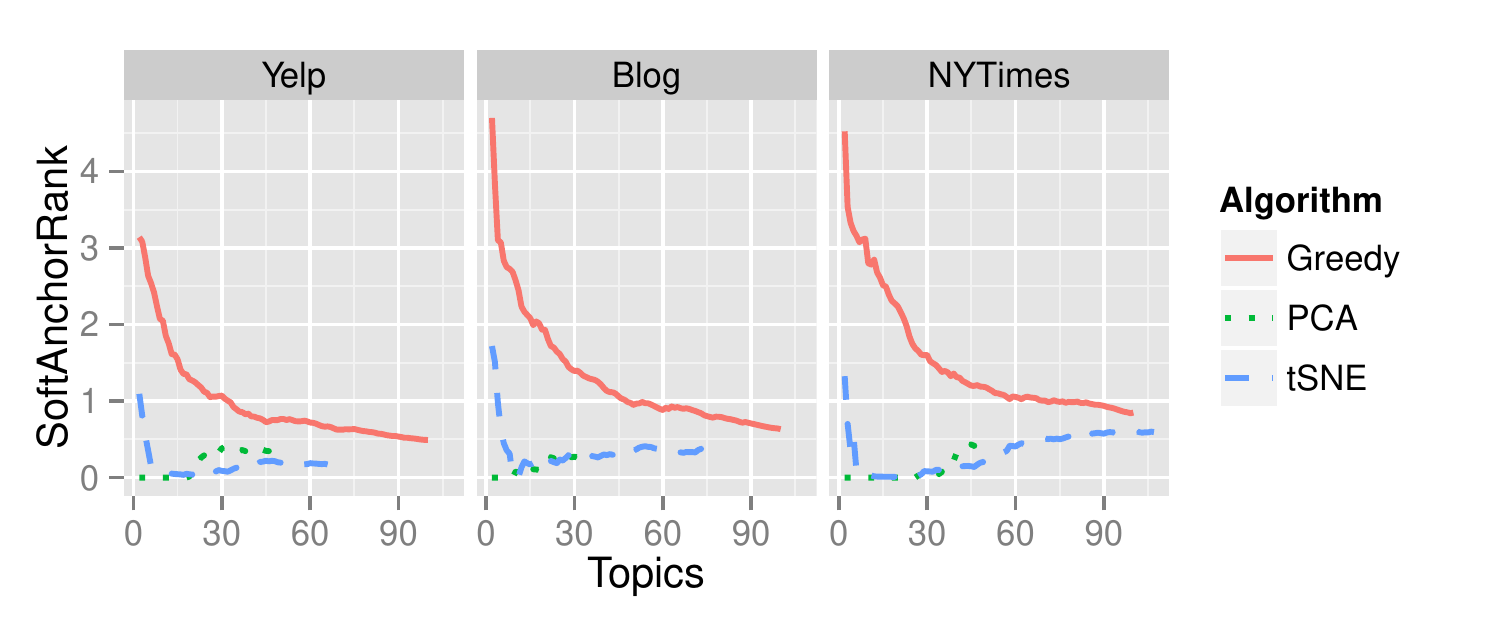}
    \caption {Anchor words have higher probability, and therefore greater salience, in $t$-SNE and PCA models (1 $\approx$ one third the probability of the top ranked word).}
    \label{fig:softrank}
\end{center}
\end{figure}
Lower values of soft rank (Fig. \ref{fig:softrank} indicate that the anchor word has greater relative probability to occur within a topic.
As we increase the number of topics, anchor words become more prominent in topics learned by the greedy method, but $t$-SNE anchor words remain relatively more probable within their topics as measured by soft rank.

\textbf{Held-out Probability}. 
Given an estimate of the topic-word matrix $A$, we can compute the marginal probability of held-out documents under that model.
We use the left-to-right estimator introduced by \cite{wallach2009evaluation}, which uses a sequential algorithm similar to a Gibbs sampler.
This method requires a smoothing parameter for document-topic Dirichlet distributions, which we set to $\alpha_k = 0.1$.
\begin{figure}[htb]
\begin{center}
\hspace*{-0.27cm}
    \includegraphics[scale=0.55]{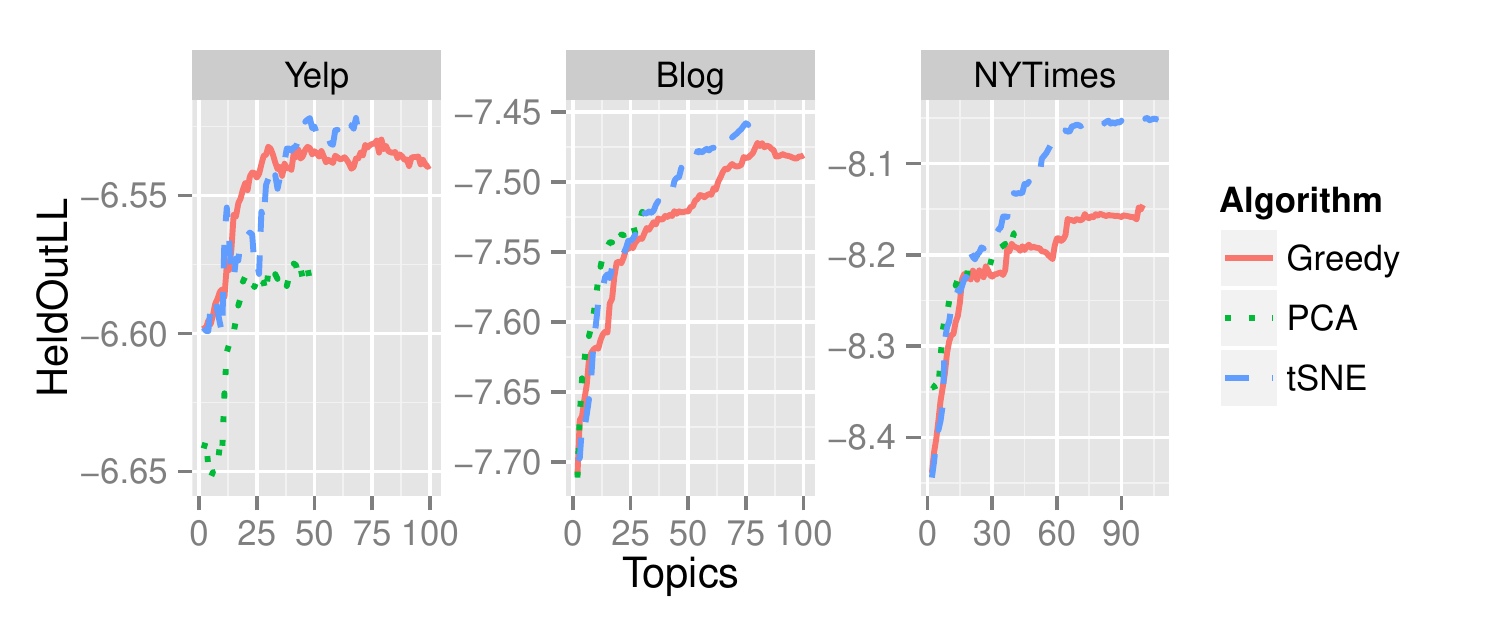}
    \caption {$t$-SNE topics have better held-out probability than greedy topics.}
    \label{fig:heldout}
\end{center}
\end{figure}
We note that the greedy algorithm run on the original, unprojected matrix has better held-out probability values than $t$-SNE for the Yelp corpus, but as this method does not scale to realistic vocabularies we compare here to the sparse random projection method used in \cite{arora2013practical}.
The $t$-SNE method appears to do best, particularly in the NYT corpus, which has a larger vocabulary and longer training documents. The length of individual held-out documents has no correlation with held-out probability.

We emphasize that Held-out Probability is sensitive to smoothing parameters and should only be used in combination with a range of other topic-quality metrics.
In initial experiments, we observed significantly worse held-out performance for the $t$-SNE algorithm.
This phenomenon was because setting the probability of anchor words to zero for all but their own topics led to large negative values in held-out log probability for those words.
As $t$-SNE tends to choose more frequent terms as anchor words, these ``spikes'' significantly affected overall probability estimates.
To make the calculation more fair, we added $10^{-5}$ to any zero entries for anchor words in the topic-word matrix $A$ across {\em all} models and renormalized.

Because $t$-SNE is a stochastic model, different initializations can result in different embeddings.
To evaluate how steady anchor word selection is, we ran five random initializations for each dataset.
For the Yelp dataset, the number of anchor words varies from 59 to 69, and 43 out of those are shared across at least four trials.
For the Blog dataset, the number of anchor words varies from 80 to 95, with 56 shared across at least four trials.
For the NYT dataset, this number varies between 83 and 107, with 51 shared across at least four models.

\subsection{Qualitative Results}

Table  \ref{tbl:examples} shows topics trained by three methods ($t$-SNE, PCA, and greedy) for all three datasets.
For each model, we select five topics {\em at random} from the $t$-SNE model, and then find the closest topic from each of the other models.
If anchor words present in the top eight words, they are shown in boldface.

A fundamental difference between anchor-based inference and traditional likelihood-based inference is that we can give an {\em order} to topics according to their contribution to word co-occurrence convex hull. 
This order is intrinsic to the original algorithm, and we heuristically give orders to $t$-SNE and PCA based on their contributions.
This order is listed as \# in the previous table.
For all but one topic, the closest topic from the greedy model has a higher order number than the associated $t$-SNE topic.
As shown above, the standard algorithm tends to pick less useful anchor words at the initial stage; only the later, higher ordered topics are specific.

The most clear distinction between models is the rank of anchor words represented by Hard Rank for each topic.
Only one topic corresponding to ({\em initial}) has the anchor word which does not coincide with the top-ranked word.
For the greedy algorithm, anchor words are often tens of words down the list in rank, indicating that they are unlikely to find a connection to the topic's semantic core.
In cases where the anchor word is highly ranked ({\em unbelievers}, {\em parenthood}) the word is a good indicator of the topic, but still less decisive. 

$t$-SNE and PCA are often consistent in their selection of anchor words, which provides useful validation that low-dimensional embeddings discern more relevant anchor words regardless of linear vs non-linear projections.
Note that we are only varying the anchor selection part of the Anchor Words algorithm in these experiments, recovering topic-word distributions in the same manner given anchor words.
As a result, any differences between topics with the same anchor word (for example {\em chicken}) are due to the difference in either the number of topics or the rest of anchor words. 
Since PCA suffers from a crowding problem in lower-dimensional projection (see Figure \ref{fig:pca-2d}) and the problem could be severe in a dataset with a large vocabulary, $t$-SNE is more likely to find the proper number of anchors given a specified granularity.

\section{Conclusion}

One of the main advantages of the anchor words algorithm is that the running time is largely independent of corpus size. 
Adding more documents would not affect the size of the co-occurrence matrix, requiring more times to construct the co-occurrence matrix at the beginning.
While the inference is scalable depending only on the size of the vocabulary, finding quality anchor words is crucial for the performance of the inference.

\cite{arora2013practical} presents a greedy anchor finding algorithm that improves over previous linear programming methods, but finding quality anchor words remains an open problem in spectral topic inference. 
We have shown that previous approaches have several limitations.
Exhaustively finding anchor words by eliminating words that are reproducible by other words \cite{AGM} is impractical.
The anchor words selected by the greedy algorithm are overly eccentric, particularly at the early stages of the algorithm, causing topics to be poorly differentiated.
We find that using low-dimensional embeddings of word co-occurrence statistics allows us to approximate a better convex hull.
The resulting anchor words are highly {\em salient}, being both distinctive and probable.
The models trained with these words have better quantitative and qualitative properties along various metrics.
Most importantly, using radically low-dimensional projections allows us to provide users with clear visual explanations for the model's anchor word selections.

An interesting property of using low-dimensional embeddings is that the number of topics depends only on the projecting dimension.
Since we can efficiently find an exact convex hull in low-dimensional space, users can achieve topics with their preferred level of granularities by changing the projection dimension.
We do not insist this is the ``correct'' number of topics for a corpus, but this method, along with the range of metrics described in this paper, provides users with additional perspective when choosing a dimensionality that is appropriate for their needs.

We find that the $t$-SNE method, besides its well-known ability to produce high quality layouts, provides the best overall anchor selection performance.
This method consistently selects higher-frequency terms as anchor words, resulting in greater clarity and interpretability.
Embeddings with PCA are also effective, but they result in less well-formed spaces, being less effective in held-out probability for sufficiently large corpora.

Anchor word finding methods based on low-dimensional projections offer several important advantages for topic model users.
In addition to producing more salient anchor words that can be used effectively as topic labels, the relationship of anchor words to a visualizable word co-occurrence space offers significant potential.
Users who can see why the algorithm chose a particular model will have greater confidence in the model and in any findings that result from topic-based analysis.
Finally, visualizable spaces offer the potential to produce interactive environments for semi-supervised topic reconstruction.

\section*{Acknowledgments}

We thank David Bindel and the anonymous reviewers for their valuable comments and suggestions,
and Laurens van der Maaten for providing his $t$-SNE implementation.

\bibliography{refs}
\bibliographystyle{acl}


\end{document}